# Reconstructing Syllable Sequences in Abugida Scripts with Incomplete Inputs

Ye Kyaw Thu[1,2][0000−0003−3115−6166] and
Thazin Myint Oo[2][0000−0002−0544−4941]

[1] Language and Semantic Technology Research Team, NECTEC, Thailand
[2] Language Understanding Laboratory, Myanmar
yekyaw.thu@nectec.or.th, queenofthazin@gmail.com

Abstract. This paper explores syllable sequence prediction in Abugida languages using Transformer-based models, focusing on six languages: Bengali, Hindi, Khmer, Lao, Myanmar, and Thai, from the Asian Language Treebank (ALT) dataset. We investigate the reconstruction of complete syllable sequences from various incomplete input types, including consonant sequences, vowel sequences, partial syllables (with random character deletions), and masked syllables (with fixed syllable deletions). Our experiments reveal that consonant sequences play a critical role in accurate syllable prediction, achieving high BLEU scores, while vowel sequences present a significantly greater challenge. The model demonstrates robust performance across tasks, particularly in handling partial and masked syllable reconstruction, with strong results for tasks involving consonant information and syllable masking. This study advances the understanding of sequence prediction for Abugida languages and provides practical insights for applications such as text prediction, spelling correction, and data augmentation in these scripts.

Keywords: Abugida Languages · Neural Machine Translation · Transformer Architecture · Syllable Reconstruction · Syllable Sequence Prediction · Text Generation · Spelling Correction · Data Augmentation

## 1 Introduction

In recent years, Transformer-based models have revolutionized the fields of artificial intelligence (AI) and natural language processing (NLP), particularly in tasks such as text generation, machine translation, and language modeling. These models, known for their ability to capture long-range dependencies and contextual information, have become the foundation of many state-of-the-art systems, including large language models (LLMs). While much of the research has focused on widely studied languages and scripts, there remains a significant gap in understanding how these models perform on languages with unique writing systems, such as Abugida scripts [3, 4].

Abugida languages, which include Bengali, Hindi, Khmer, Lao, Myanmar, Thai, and others, share a common characteristic in their syllable-based writing systems. In these scripts, syllables are typically formed by combining consonants with inherent vowels, which can be modified or replaced by diacritics or additional vowel markers. This unique structure presents both challenges and opportunities for NLP tasks, particularly in scenarios involving incomplete or partially observed text inputs.

In this study, we explore the application of Transformer-based models to the task of syllable sequence prediction in Abugida scripts. Specifically, we investigate the reconstruction of complete syllable sequences from various types of incomplete inputs, including:

- Consonant sequences (e.g., sequences containing only consonants from each syllable),
- Vowel sequences (e.g., sequences containing only vowels from each syllable),
- Partial syllable sequences (e.g., syllables with one or two characters randomly deleted), and
- Masked syllable sequences (e.g., sequences with a fixed number of syllables masked, such as 3, 5, 8, or 10 syllables).

We focus on six Abugida languages, namely Bengali, Hindi, Khmer, Lao, Myanmar, and Thai, using data from the Asian Language Treebank (ALT) dataset [1, 2]. Our experiments reveal several key findings:

- Consonant sequences play a critical role in accurate syllable prediction, achieving high BLEU scores across all languages. For example, the model achieves BLEU scores ranging from 62.65 to 99.87 for consonant-to-syllable prediction tasks.
- Vowel sequences, on the other hand, present a significantly greater challenge, with BLEU scores ranging from 13.62 to 60.28, highlighting the importance of consonant information in syllable reconstruction.
- The model demonstrates strong performance in handling partial syllable reconstruction, achieving BLEU scores ranging from 65.19 to 88.59 for 1-random-deletion tasks and 75.24 to 99.10 for 2-random-deletion tasks.
- Additionally, the model performs exceptionally well on syllable masking tasks, achieving BLEU scores above 91.51 for 3-syllable masking.

These results underscore the robustness of Transformer-based models in handling incomplete text inputs and their potential for applications such as text prediction, spelling correction, and data augmentation in Abugida scripts. The findings also advance our understanding of sequence prediction for languages with unique writing systems, providing practical insights for improving language technologies in these scripts.

For detailed information on the experimental settings, methodology, and results, please refer to the respective sections of this paper.

## 2 Abugida Languages

Abugida languages are characterized by a writing system that combines features of both syllabic and segmental scripts. Unlike alphabetic systems, where each character typically represents a single phoneme (e.g., Latin script), or purely syllabic systems, where each character represents a full syllable (e.g., Japanese hiragana), abugidas use a base consonant that inherently carries a default vowel sound. This vowel can be modified or replaced by diacritics or additional vowel markers attached to the consonant. This unique structure allows abugidas to efficiently represent complex syllable structures while maintaining a relatively compact script. Examples of abugida scripts include those used for Bengali, Hindi, Khmer, Lao, Myanmar, and Thai, which are the focus of this study.

The abugida writing system is widely used across South and Southeast Asia, with each language adapting the script to its phonological and grammatical requirements. For instance, in Khmer, the script includes numerous diacritics and subscript consonants to represent complex syllable structures, while in Thai, tone markers are integrated into the script to convey tonal distinctions. Despite these variations, all abugida scripts share the common feature of combining consonants with inherent vowels, which can be altered by diacritics. This shared structure makes abugida languages particularly interesting for computational studies, as the scripts present unique challenges for tasks such as text prediction, spelling correction, and data augmentation.

The complexity of abugida scripts, particularly the use of diacritics and the visual similarity of glyphs, poses significant challenges for natural language processing (NLP) systems. For example, minor changes in diacritics can alter the meaning of a word, and visually similar glyphs (homoglyphs) can lead to ambiguity in text processing. These challenges are further compounded in low-resource settings, where limited data availability makes it difficult to train robust models. Understanding and addressing these challenges is crucial for developing effective NLP tools for abugida languages, which are spoken by hundreds of millions of people worldwide. For a detailed illustration of how syllables are constructed in the six Abugida languages studied here, see Table 1.

Table 1: Syllable Construction with Common Vowel Sounds in Abugida Languages

| Language | a | ā | i | ī | u |
|---|---|---|---|---|---|
| Bengali (bn) | ক (ka) | কা (kā) | কি (ki) | কী (kī) | কু (ku) |
| Hindi (hi) | क (ka) | का (kā) | कि (ki) | की (kī) | कु (ku) |
| Khmer (km) | ក (ka) | កា (kā) | កិ (ki) | កី (kī) | កុ (ku) |
| Lao (lo) | ກ (ka) | ກາ (kā) | ກິ (ki) | ກີ (kī) | ກຸ (ku) |
| Myanmar (my) | က (ka) | ကာ (kā) | ကိ (ki) | ကီ (kī) | ကု (ku) |
| Thai (th) | ก (ka) | กา (kā) | กิ (ki) | กี (kī) | กุ (ku) |

## 3 Data Preparation

The dataset used in this study is derived from the Asian Language Treebank (ALT), a parallel corpus developed to advance natural language processing (NLP) research for Asian languages [1, 2]. The ALT corpus consists of approximately 20,000 sentences sampled from English Wikinews, which were manually translated into six Asian languages: Bengali, Hindi, Khmer, Lao, Myanmar, and Thai. Each sentence is annotated with word segmentation, part-of-speech (POS) tags, syntactic trees, and word alignment links to the English source text.

For our experiments, we focused on the six Abugida languages from the ALT corpus: Bengali (bg), Hindi (hi), Khmer (kh), Lao (lo), Myanmar (my), and Thai (th). These languages were chosen due to their unique syllable-based writing systems, which present interesting challenges for sequence prediction tasks.

### 3.1 Data Cleaning

The original ALT dataset underwent two cleaning steps to prepare it for our experiments:

First Cleaning: In this step, we removed all non-language-specific characters (e.g., punctuation, numbers, and symbols) using a Python script. The script retained only valid Unicode characters for each language, ensuring that the dataset contained only linguistically relevant text. For example, for Bengali, the script retained characters within the Unicode range [U0980-U09FF].

Second Cleaning: The second cleaning step focused on removing unwanted characters such as independent vowels, numbers, and additional punctuation marks that were not relevant to our study. This step ensured that the dataset was further refined for syllable segmentation and subsequent tasks.

### 3.2 Syllable Segmentation

After cleaning, syllable segmentation was performed for all six languages using a Python script. The segmentation process involved breaking down words into their constituent syllables based on language-specific rules. For example, in Myanmar, syllables are typically formed by combining consonants with inherent vowels, which can be modified by diacritics. The segmentation script used regular expressions tailored to each language's writing system to ensure accurate syllable boundaries.

### 3.3 Data Statistics

The final dataset consists of syllable-segmented sentences for each language, along with extracted consonant and vowel sequences. Table 2 provides an overview of the dataset statistics, including the number of sentences, syllables, consonants, vowels, and characters for each language.

The table highlights the linguistic diversity of the dataset, with Khmer having the highest number of syllables and characters, while Myanmar has the highest number of vowels due to its complex diacritic system. These statistics provide a foundation for understanding the dataset's composition and its suitability for sequence prediction tasks.

Table 2: Dataset Statistics for Six Abugida Languages

| Language | Sentences | Syllables | Consonants | Vowels |
|---|---|---|---|---|
| Bengali (bg) | 20,104 | 1,305,758 | 1,299,812 | 689,980 |
| Hindi (hi) | 20,104 | 1,204,982 | 1,199,923 | 668,981 |
| Khmer (kh) | 20,104 | 1,623,075 | 1,620,448 | 619,771 |
| Lao (lo) | 20,104 | 1,305,103 | 1,305,073 | 508,609 |
| Myanmar (my) | 20,104 | 1,061,973 | 1,059,278 | 913,284 |
| Thai (th) | 20,104 | 1,364,769 | 1,364,769 | 551,146 |

### 3.4 Random Character Deletion

To simulate incomplete or noisy input scenarios, we applied a random character deletion process to the syllable-segmented data for all six Abugida languages. This step involved randomly removing one or two characters from each syllable in the dataset. The goal was to create partially observed input sequences that mimic real-world scenarios where text data may be incomplete or corrupted.

### 3.5 Masking Syllables

As part of the data preparation for syllable sequence prediction, syllable masking was conducted on all six Abugida languages. For each language, syllables were masked using `-mask` values of 3, 5, 8, and 10, ensuring that no two masked syllables were consecutive. The results, summarized in Table 3, show high masking success rates across all languages, with 99.9% of sentences successfully masked for `-mask 3` and `-mask 5`. As the `-mask` value increased, the percentage of skipped sentences (due to insufficient syllables) remained very low ($\leq$ 0.10%). Khmer exhibited the highest average syllables per sentence (81.06), while Myanmar had the lowest (53.03), resulting in slightly more skipped sentences for higher `-mask` values. The total number of syllables masked increased proportionally with the `-mask` value, with Bengali and Thai showing the highest totals (e.g., 180,641 and 180,470 syllables masked, respectively, for `-mask 10`).

### 3.6 Processed Data Examples

Figures 1 and 2 illustrate examples of processed data for Myanmar and Hindi sentences, respectively. Each figure demonstrates the original syllable-segmented sentence, followed by derived sequences such as consonant-only and vowel-only representations. Additionally, the figures showcase variations of the input data, including sentences with one or two characters randomly deleted from each syllable, as well as sentences with 10 randomly masked syllables (replaced with underscores). These examples highlight the diverse input variations used in our experiments, which are designed to simulate incomplete or noisy text scenarios for robust sequence prediction tasks.

Table 3: Summary of Masking Statistics for Each Language and Mask Value

| Language | Mask Value | Masked | Skipped (%) | Total Syl Masked |
|---|---|---|---|---|
| Bengali | 3 | 20,104 | 0.00 | 60,312 |
| (64.95) | 5 | 20,102 | 0.01 | 100,510 |
| | 8 | 20,101 | 0.02 | 160,808 |
| | 10 | 20,100 | 0.02 | 201,000 |
| Hindi | 3 | 20,104 | 0.00 | 60,312 |
| (59.85) | 5 | 20,103 | 0.01 | 100,515 |
| | 8 | 20,100 | 0.02 | 160,800 |
| | 10 | 20,097 | 0.04 | 200,970 |
| Khmer | 3 | 20,104 | 0.00 | 60,312 |
| (80.45) | 5 | 20,103 | 0.01 | 100,515 |
| | 8 | 20,101 | 0.02 | 160,808 |
| | 10 | 20,099 | 0.03 | 200,990 |
| Lao | 3 | 20,104 | 0.00 | 60,312 |
| (64.95) | 5 | 20,104 | 0.00 | 100,520 |
| | 8 | 20,100 | 0.02 | 160,800 |
| | 10 | 20,097 | 0.04 | 200,970 |
| Myanmar | 3 | 20,096 | 0.04 | 60,288 |
| (52.85) | 5 | 20,096 | 0.04 | 100,480 |
| | 8 | 20,091 | 0.07 | 160,728 |
| | 10 | 20,086 | 0.10 | 200,860 |
| Thai | 3 | 20,104 | 0.00 | 60,312 |
| (68.15) | 5 | 20,102 | 0.01 | 100,510 |
| | 8 | 20,100 | 0.02 | 160,800 |
| | 10 | 20,095 | 0.05 | 200,950 |

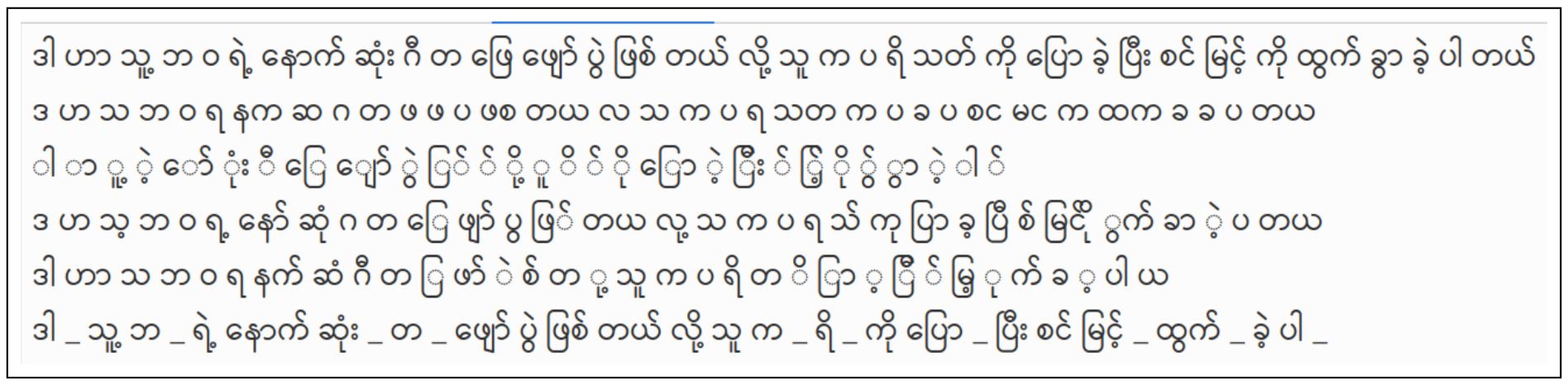

Fig. 1: Examples of processed data for a Myanmar sentence. From top to bottom, the figure shows: the original sentence (cleaned by removing punctuation and non-Myanmar characters, and segmented into syllables); the consonant-only sequence; the vowel-only sequence; a version with one character randomly deleted from each syllable; a version with two characters randomly deleted from each syllable; and a version where 10 syllables are randomly masked with underscores.

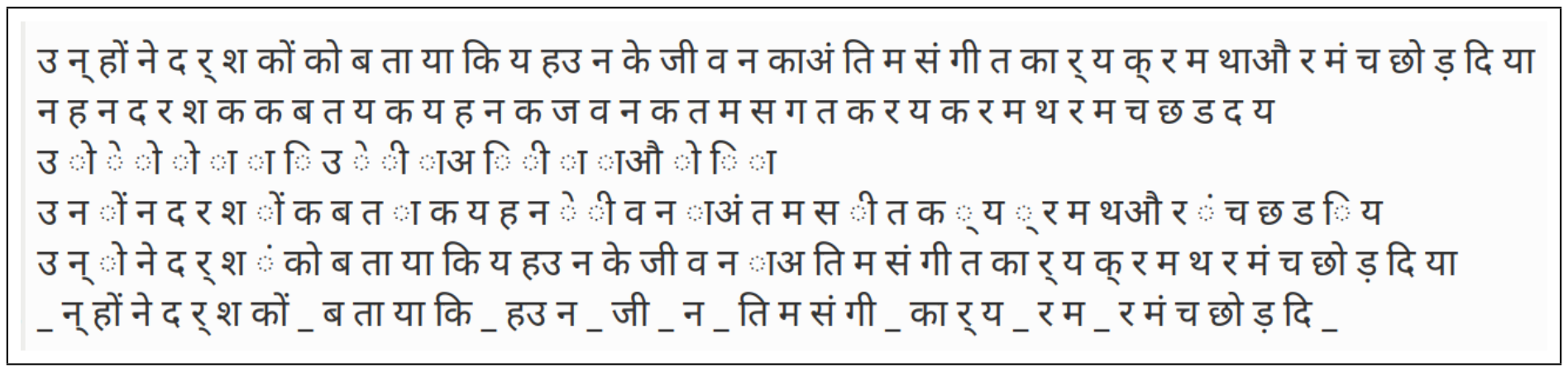

Fig. 2: Examples of processed data for a Hindi sentence. From top to bottom, the figure shows: the original sentence (cleaned by removing punctuation and non-Hindi characters, and segmented into syllables); the consonant-only sequence; the vowel-only sequence; a version with one character randomly deleted from each syllable; a version with two characters randomly deleted from each syllable; and a version where 10 syllables are randomly masked with underscores.

## 4 Experimental Setup

### 4.1 Parallel Data Information

For our experiments, we used parallel datasets consisting of syllable-segmented sentences for six Abugida languages. The target language data, which represents the complete syllable sequences, comprises 18,104 sentences for training, 1,000 sentences for development, and 1,000 sentences for testing. The syllable units for the target data are as follows: the training set contains 1,180,989 syllables, the development set contains 61,834 syllables, and the test set contains 62,935 syllables. For the source data, which consists of incomplete syllable sequences (e.g., consonant-only or vowel-only sequences), the number of words or segmented units varies depending on the type of input (e.g., consonant sequences, vowel sequences, or partially masked syllables). Despite this variability, the sentence-level parallelism is maintained across all datasets, ensuring consistent alignment between source and target sequences.

### 4.2 Marian NMT System

For our experiments, we utilized Marian [6], an efficient and self-contained Neural Machine Translation (NMT) framework written in C++. Marian is designed for high performance and flexibility, offering features such as dynamic computation graphs, automatic differentiation, and support for various model architectures, including Transformer-based models. We employed Marian to train and evaluate models for syllable sequence prediction tasks, using a Transformer architecture with the following hyperparameters: 2 encoder and decoder layers, 8 attention heads, dropout rate of 0.3, label smoothing of 0.1, and a learning rate of 0.0003 with inverse square root decay. The training was conducted on a single NVIDIA GeForce RTX 3090 Ti GPU with 24 GB of memory, using a beam size of 6 for decoding. Below is the shell script used for training the model, which can be adapted for further studies or reimplementation:

Listing 1.1: Shell script for training Marian NMT model.

```
#!/bin/bash

model_folder="co-my/model.tf.co2my";
mkdir -p ${model_folder};
data_path="/home/ye/exp/nmt/cv-predict/data/co/co-my";
src="co"; tgt="sy";

marian \
    --model ${model_folder}/model.npz --type transformer \
    --train-sets ${data_path}/train.${src} ${data_path}/train.${tgt} \
    --max-length 200 \
    --vocabs ${data_path}/vocab/vocab.${src}.yml \
    ${data_path}/vocab/vocab.${tgt}.yml \
    --mini-batch-fit -w 1000 --maxi-batch 100 \
    --early-stopping 10 \
    --valid-freq 5000 --save-freq 5000 --disp-freq 500 \
    --valid-metrics cross-entropy perplexity bleu \
    --valid-sets ${data_path}/dev.${src} ${data_path}/dev.${tgt} \
    --valid-translation-output ${model_folder}/valid.${src}-${tgt}.output \
```

```
    --quiet-translation \
    --valid-mini-batch 64 \
    --beam-size 6 --normalize 0.6 \
    --log ${model_folder}/train.log \
    --valid-log ${model_folder}/valid.log \
    --enc-depth 2 --dec-depth 2 \
    --transformer-heads 8 \
    --transformer-postprocess-emb d \
    --transformer-postprocess dan \
    --transformer-dropout 0.3 --label-smoothing 0.1 \
    --learn-rate 0.0003 --lr-warmup 0 \
    --lr-decay-inv-sqrt 16000 --lr-report \
    --clip-norm 5 \
    --tied-embeddings \
    --devices 0 --sync-sgd --seed 1111 \
    --exponential-smoothing \
    --dump-config > ${model_folder}/${src}-${tgt}.config.yml

time marian -c ${model_folder}/${src}-${tgt}.config.yml 2>&1 | \
tee ${model_folder}/tf.${src}-${tgt}.log
```

Marian is distributed under the MIT license and is available at `https://github.com/marian-nmt/marian`.

### 4.3 Evaluation Metrics

For evaluating the performance of our syllable sequence prediction models, we employed the BLEU (Bilingual Evaluation Understudy) score [5] as the primary metric. BLEU is a widely used automatic evaluation metric for machine translation tasks, designed to measure the similarity between a candidate translation and one or more reference translations. It computes a weighted average of modified $n$-gram precision scores, combined with a brevity penalty to penalize overly short translations. The BLEU score ranges from 0 to 1, where higher scores indicate better translation quality. A score of 1 is achieved only when the candidate translation exactly matches one of the reference translations.

During training, we monitored multiple metrics, including cross-entropy, perplexity, and BLEU, using the Marian NMT framework's `-valid-metrics` option. Cross-entropy measures the difference between the predicted probability distribution and the true distribution, providing insight into the model's learning progress. Perplexity, derived from cross-entropy, quantifies how well the model predicts the next token in a sequence, with lower values indicating better performance. While these metrics are useful for tracking training progress, they are less interpretable for final evaluation compared to BLEU.

For testing and evaluation, we focused solely on the BLEU score due to its strong correlation with human judgment and its widespread adoption in the machine translation community.

# 5 Result and Discussion

## 5.1 Analysis of BLEU Scores for Consonant and Vowel Sequences

Table 4 presents the BLEU scores for syllable sequence prediction tasks using Transformer-based neural machine translation. The results reveal a clear distinction between the performance of consonant-based and vowel-based predictions. For all six languages, the BLEU scores for consonant-to-syllable predictions are significantly higher than those for vowel-to-syllable predictions. For instance, Khmer, Lao, and Thai achieve near-perfect scores (99.38, 99.70, and 99.87, respectively) for consonant-based predictions, while their vowel-based predictions yield much lower scores (23.50, 13.62, and 33.78, respectively). This disparity underscores the critical role of consonants in syllable formation in Abugida scripts, where consonants often carry inherent vowel sounds and form the structural backbone of syllables. In contrast, vowels alone provide insufficient information for accurate syllable prediction, as they are typically modifiers or secondary components in these writing systems.

Myanmar stands out as an exception, with a relatively high vowel-based BLEU score of 60.28, suggesting that vowels play a more prominent role in its syllable structure compared to other languages. This could be attributed to Myanmar's complex diacritic system, where vowels are more explicitly represented and carry significant phonetic information. Overall, the results highlight the importance of consonants in Abugida scripts and the challenges associated with vowel-only predictions.

## 5.2 Analysis of BLEU Scores for Partial Syllable Inputs

Table 5 presents the BLEU scores for syllable sequence prediction tasks using partial syllable inputs. The results demonstrate the model's robustness in handling incomplete syllable data. For most languages, the BLEU scores improve significantly when two characters are deleted compared to one character deleted. For example, in Lao, the BLEU score increases from 76.72 (one character deleted) to 99.10 (two characters deleted), indicating that the model performs exceptionally well even when more information is missing. This suggests that the Transformer-based model can effectively infer missing characters by leveraging contextual information and the inherent structure of Abugida scripts.

However, Myanmar exhibits a different trend, with a lower BLEU score for two-character deletions (75.24) compared to one-character deletions (85.02). This could be due to the complexity of Myanmar's script, where the deletion of two characters may remove critical diacritics or modifiers, making it harder for the model to reconstruct the original syllable. The results for Khmer and Thai are particularly impressive, with BLEU scores above 97 for two-character deletions, indicating their scripts' regularity and the model's ability to handle partial inputs effectively.

### 5.3 Analysis of BLEU Scores for Randomly Masked Syllables

Table 6 presents the BLEU scores for syllable sequence prediction tasks with randomly masked syllables. The results show a consistent decline in performance as the number of masked syllables increases, reflecting the increased difficulty of the task. For example, in Bengali, the BLEU score drops from 94.25 (3 syllables masked) to 85.73 (10 syllables masked), while in Khmer, the score decreases from 97.24 to 93.63. This trend is consistent across all languages, indicating that the model's ability to predict missing syllables diminishes as the amount of missing information grows.

Despite this decline, the model maintains relatively high BLEU scores even with 10 syllables masked, demonstrating its robustness in handling incomplete inputs. Khmer and Thai consistently achieve the highest scores across all masking levels, likely due to the regularity and predictability of their scripts. In contrast, Myanmar exhibits the lowest scores, particularly for higher masking levels (e.g., 78.74 for 10 syllables masked), further highlighting the challenges posed by its complex diacritic system.

### 5.4 Comparative Discussion

The results from all three tables collectively highlight the Transformer-based model's ability to handle various types of incomplete inputs for syllable sequence prediction in Abugida scripts. Consonant-based predictions consistently outperform vowel-based predictions, emphasizing the structural importance of consonants in these scripts. The model's strong performance on partial syllable inputs, particularly for two-character deletions, demonstrates its ability to infer missing information effectively. However, the performance varies across languages, with Khmer and Thai achieving the highest scores due to their regular and predictable scripts, while Myanmar's complex diacritic system poses greater challenges.

The decline in performance with increasing numbers of masked syllables underscores the importance of contextual information in syllable prediction. While the model performs well with a small number of masked syllables, its ability to reconstruct sequences diminishes as more information is removed. This suggests that future work could focus on enhancing the model's ability to handle larger gaps in input data, potentially by incorporating additional linguistic features or leveraging larger training datasets.

## 6 Conclusion

This study explored the application of Transformer-based models to syllable sequence prediction in six Abugida languages (Bengali, Hindi, Khmer, Lao, Myanmar, and Thai) using the Asian Language Treebank (ALT) dataset. We investigated the reconstruction of complete syllable sequences from various incomplete input types, including consonant sequences, vowel sequences, partial syllables (with random character deletions), and masked syllables (with fixed syllable deletions). Our experiments yielded several key findings:

Table 4: BLEU Scores for Syllable Sequence Prediction from Consonant and Vowel Sequences

| Language | Consonant to Syllables | Vowel to Syllables |
|---|---|---|
| Bangladeshi (bg) | 67.24 | 16.74 |
| Hindi (hi) | 62.65 | 22.69 |
| Khmer (kh) | 99.38 | 23.50 |
| Lao (lo) | 99.70 | 13.62 |
| Myanmar (my) | 75.01 | 60.28 |
| Thai (th) | 99.87 | 33.78 |

Table 5: BLEU Scores for Syllable Sequence Prediction from Partial Syllables

| Language | One Character Deleted | Two Characters Deleted |
|---|---|---|
| Bangladeshi (bg) | 65.19 | 94.27 |
| Hindi (hi) | 67.60 | 92.33 |
| Khmer (kh) | 83.32 | 97.91 |
| Lao (lo) | 76.72 | 99.10 |
| Myanmar (my) | 85.02 | 75.24 |
| Thai (th) | 88.59 | 97.39 |

Table 6: BLEU Scores for Syllable Sequence Prediction with Randomly Masked Syllables

| Language | 3 Syllables Masked | 5 Syllables Masked | 8 Syllables Masked | 10 Syllables Masked |
|---|---|---|---|---|
| Bengali (bg) | 94.25 | 91.68 | 88.16 | 85.73 |
| Hindi (hi) | 93.36 | 90.51 | 86.56 | 84.16 |
| Khmer (kh) | 97.24 | 96.15 | 94.52 | 93.63 |
| Lao (lo) | 94.52 | 91.66 | 87.83 | 85.43 |
| Myanmar (my) | 91.51 | 87.39 | 81.77 | 78.74 |
| Thai (th) | 96.32 | 94.91 | 92.77 | 91.37 |

1. Consonant Sequences Are Critical: Consonant-based predictions consistently achieved high BLEU scores, highlighting the structural importance of consonants in Abugida scripts. For example, Khmer, Lao, and Thai achieved near-perfect scores (above 99) for consonant-to-syllable predictions, underscoring the robustness of the model in leveraging consonant information.

2. Vowel Sequences Pose Challenges: Vowel-based predictions yielded significantly lower BLEU scores, with Myanmar being the only exception due to its complex diacritic system. This emphasizes the limited role of vowels in syllable reconstruction and the need for consonant information in Abugida scripts.

3. Robustness to Partial Inputs: The model demonstrated strong performance in handling partial syllable inputs, particularly for two-character deletions. This suggests that the Transformer-based architecture effectively leverages contextual information to infer missing characters, even in the presence of incomplete data.

4. Handling Masked Syllables: While the model maintained high BLEU scores for tasks involving a small number of masked syllables, its performance declined as the number of masked syllables increased. This highlights the importance of contextual information and the challenges of reconstructing sequences with larger gaps.

These findings advance our understanding of sequence prediction in Abugida scripts and demonstrate the potential of Transformer-based models for applications such as text prediction, spelling correction, and data augmentation in low-resource languages. However, this study has certain limitations. First, the experiments were conducted on a relatively small corpus of approximately 20,000 sentences from the news domain, which may limit the generalizability of the results to other domains. Second, the study did not explore syllable sequence prediction with specific syllable formation information, such as left vowels, right vowels, or stacked consonants, which are critical features of Abugida scripts. Future work could focus on fine-tuning approaches using Large Language Models (LLMs) to improve the reconstruction of syllable sequences, as well as expanding the dataset to include more diverse domains and languages. To facilitate reproducibility and further research, we have made all configuration files, code, and preprocessed data publicly available at `https://github.com/ye-kyaw-thu/preprints/tree/main/2_Syllable_Reconstruct`.